\documentclass[journal]{IEEEtran}

\usepackage{amsmath,amsfonts}
\usepackage{dsfont}
\usepackage{algorithmic}
\usepackage{algorithm}
\usepackage{array}
\usepackage[caption=false,font=normalsize,labelfont=sf,textfont=sf]{subfig}
\usepackage{textcomp}
\usepackage{multirow}
\usepackage{stfloats}
\usepackage{url}
\usepackage{verbatim}
\usepackage{graphicx}
\usepackage{booktabs}
\usepackage[square,sort&compress,numbers]{natbib}
\usepackage{pifont}
\usepackage[hidelinks]{hyperref}

\begin{document}

\title{CloudMamba: An Uncertainty-Guided Dual-Scale Mamba Network for Cloud Detection in Remote Sensing Imagery}

\author{Jiajun Yang, Keyan Chen, Zhengxia Zou,~\IEEEmembership{Senior Member,~IEEE}, and Zhenwei Shi$^\star$,~\IEEEmembership{Senior Member,~IEEE}

\thanks{This work was supported in part by the National Natural Science Foundation of China under Grants 62125102, U24B20177, U25A20401 and 62471014, in part by the Inner Mongolia Autonomous Region Science and Technology Planning Project under Grant 2025YFHH0124, and in part by the Fundamental Research Funds for the Central Universities. \emph{(Corresponding Author: Zhenwei Shi (shizhenwei@buaa.edu.cn))}}

\thanks{Jiajun Yang, Keyan Chen, Zhengxia Zou, and Zhenwei Shi are with the Department of Aerospace Intelligent Science and Technology, School of Astronautics, Beihang University, Beijing 100191, China, and also with the Key Laboratory of Spacecraft Design Optimization and Dynamic Simulation Technologies, Ministry of Education, Beihang University, Beijing 100191, China.}
}

\markboth{Journal of \LaTeX\ Class Files,~Vol.~XX, No.~XX, XX~2026}%
{Shell \MakeLowercase{\textit{et al.}}: A Sample Article Using IEEEtran.cls for IEEE Journals}


\maketitle

\begin{abstract}
Cloud detection in remote sensing imagery is a fundamental, critical, and highly challenging problem. Existing deep learning-based cloud detection methods generally formulate it as a single-stage pixel-wise binary segmentation task with one forward pass. However, such single-stage approaches exhibit ambiguity and uncertainty in thin-cloud regions and struggle to accurately handle fragmented clouds and boundary details. In this paper, we propose a novel deep learning framework termed CloudMamba. To address the ambiguity in thin-cloud regions, we introduce an uncertainty-guided two-stage cloud detection strategy. An embedded uncertainty estimation module is proposed to automatically quantify the confidence of thin-cloud segmentation, and a second-stage refinement segmentation is introduced to improve the accuracy in low-confidence hard regions. To better handle fragmented clouds and fine-grained boundary details, we design a dual-scale Mamba network based on a CNN-Mamba hybrid architecture. Compared with Transformer-based models with quadratic computational complexity, the proposed method maintains linear computational complexity while effectively capturing both large-scale structural characteristics and small-scale boundary details of clouds, enabling accurate delineation of overall cloud morphology and precise boundary segmentation. Extensive experiments conducted on the GF1\_WHU and Levir\_CS public datasets demonstrate that the proposed method outperforms existing approaches across multiple segmentation accuracy metrics, while offering high efficiency and process transparency. Our code is available at \url{https://github.com/jayoungo/CloudMamba}.
\end{abstract}

\begin{IEEEkeywords}
% Article submission, IEEE, paper, typesetting.
Remote sensing, cloud detection, state space model, uncertainty estimation
\end{IEEEkeywords}

\section{Introduction}
\IEEEPARstart{R}{emote} sensing technology, as a critical means of acquiring surface information, has played an indispensable role in numerous fields, including resource monitoring, urban planning, ecological investigation, and disaster early warning~\cite{lv2022spatial,gao2025adaptive,liu2025remote,xu2017multisource,li2026agrifm,fan2022fine,zhang2024bifa,eddin2023location,li2026cnn}. However, clouds inevitably appear in remotely sensed imagery during the acquisition process. On the one hand, clouds contain abundant meteorological and climatological information, serving as essential data sources for atmospheric analysis and cloud physics research. On the other hand, cloud coverage occludes ground objects, weakens their texture and spectral characteristics, and affects radiometric consistency and geometric registration accuracy, thereby significantly interfering with downstream tasks such as land-use analysis, remote sensing image classification, change detection, and object recognition~\cite{zhang2025nirnet,gao2025msfmamba,chen2026rsrefseg,liu2023decoupling,li2025fine,susskind2003retrieval,zhang2025foba,yang2024structural,chen2025seg}. Therefore, automatic, reliable, and fine-grained detection and segmentation of cloud regions constitute a key step in remote sensing image preprocessing and a fundamental task for ensuring data quality control and application reliability.

Extensive efforts have been devoted to cloud detection. Traditional machine learning-based methods combine handcrafted features—such as texture, color, and gradient—with classifiers (e.g., random forests and support vector machines) to enhance discrimination among different cloud types~\cite{kang2018coarse,fu2018cloud,chen2007support,sui2019energy,an2015scene,yuan2015bag,tan2016cloud,deng2018cloud}. Nevertheless, these methods are constrained by limited feature representation capability and insufficient generalization performance. In recent years, with the rapid development of deep learning, convolutional neural networks (CNNs) and encoder-decoder semantic segmentation architectures have become the mainstream paradigm for cloud detection. End-to-end segmentation frameworks can automatically learn joint spectral-spatial features and have achieved substantial improvements in overall detection accuracy. Transformer-based approaches further enhance long-range dependency modeling~\cite{mateo2017convolutional,xie2017multilevel,long2015fully,wu2018utilizing,yan2018cloud,ronneberger2015u,francis2019cloudfcn,wieland2019multi,jeppesen2019cloud,yang2019cdnet,yu2020effective,li2020deep,wu2021geographic,yang2024weakly,zhang2021improving,guo2020cloud,liu2022transcloudseg,zhang2022cloudformer,zhang2022cloudvit,ge2024cd,gong2023hybrid}. Despite these advances, existing methods still exhibit notable limitations in fine-grained cloud segmentation and uncertainty modeling in ambiguous regions.

Remote sensing cloud detection is a highly complex and challenging problem, whose core difficulties are mainly reflected in the following three interrelated aspects. First, insufficient uncertainty modeling: thin-cloud regions exhibit significant overlap with background surfaces in terms of spectral, textural, and structural characteristics, and their boundaries often present gradual transitions or even local mixing, leading to ambiguous predictions with high uncertainty; however, most existing methods formulate cloud detection as a single-stage pixel-wise binary segmentation problem, lacking explicit modeling and feedback refinement mechanisms for uncertain regions, which results in frequent false positives or false negatives in challenging areas such as thin clouds, cloud boundaries, and highly reflective surfaces. Second, limited capability in multi-scale and complex structure representation: clouds in real-world scenes are typically fragmented, with significant scale variations and complex morphologies, and cloud boundary regions contain rich and irregular geometric details, posing higher demands on the model's ability to simultaneously capture global structures and local details. Finally, limitations of modeling paradigms: convolutional architectures are biased toward local receptive fields and struggle to fully capture the global spatial dependencies of clouds; although Transformers possess strong global modeling capabilities, their quadratic computational complexity and high memory consumption restrict their application in high-resolution remote sensing scenarios, and they also face challenges in achieving an optimal balance between global structure modeling and fine-grained boundary delineation.

To address these challenges, we propose a novel cloud detection framework for remote sensing imagery, termed CloudMamba, which enhances fine-grained segmentation accuracy and robustness from two perspectives: uncertainty modeling and multi-scale structural representation. First, to mitigate prediction ambiguity and unstable confidence in thin-cloud regions, we design an uncertainty-guided two-stage cloud detection framework. After the first-stage prediction, an embedded uncertainty estimation module evaluates pixel-wise confidence to automatically identify potential thin-cloud and ambiguous boundary regions. Low-confidence regions are then forwarded to a second-stage refinement network, forming a progressive inference paradigm of "global detection--local focused refinement". This mechanism explicitly emphasizes hard regions and effectively reduces thin-cloud misclassification and boundary ambiguity. Second, to jointly model large-scale cloud structures and fine geometric boundary details, we construct a dual-scale Mamba architecture with a CNN-Mamba hybrid design. The Mamba structure achieves strong sequence modeling and long-range dependency representation under linear computational complexity. Through dual-scale modeling at large receptive fields and fine-resolution levels, the network simultaneously captures long-range associations and boundary details, achieving a favorable balance between large-scale cloud morphology and small-scale boundary precision. Compared with traditional Transformer-based models, the proposed architecture offers superior efficiency and structural adaptiveness.

Extensive experiments are conducted on the GF1\_WHU~\cite{li2017multi} and Levir\_CS~\cite{wu2021geographic} public datasets, including comprehensive comparisons with multiple deep learning-based cloud detection methods, ablation studies, and module effectiveness analyses. Experimental results demonstrate that CloudMamba achieves significant improvements in overall cloud detection accuracy, uncertainty modeling, boundary consistency, and robustness in fragmented cloud regions. Further visualization and uncertainty evaluation results indicate that the proposed two-stage refinement mechanism effectively reduces misclassification risks in low-confidence areas, while the dual-scale Mamba structure plays a critical role in enhancing multi-scale representation and cross-region structural modeling.

The main contributions of this work are summarized as follows:

(1) An uncertainty-guided two-stage cloud detection framework is proposed, which explicitly models and corrects low-confidence regions, effectively alleviating segmentation ambiguity in thin-cloud and blurred boundary areas and improving reliability and stability.

(2) A dual-scale CNN-Mamba hybrid architecture is designed to achieve collaborative representation of global cloud structures and local boundary details under linear computational complexity, enhancing fine-grained segmentation capability for complex cloud morphologies and fragmented structures.

(3) Extensive experiments and ablation studies on multiple public cloud detection datasets demonstrate that CloudMamba outperforms existing methods in terms of accuracy and robustness, providing a structurally transparent and interpretable paradigm for remote sensing cloud detection.

\section{Related Work}
\subsection{Traditional Machine Learning-Based Cloud Detection in Remote Sensing Images}
Early cloud detection methods mainly rely on the spectral characteristics of remote sensing imagery and human prior knowledge, distinguishing clouds from background surfaces through brightness thresholds, color space transformations, cloud indices, or empirical rules. These methods are simple in structure and highly interpretable, but they typically depend on manually defined thresholds and are sensitive to sensor types, land cover, and imaging conditions. To improve generalization and robustness, researchers have introduced machine learning methods such as Support Vector Machine (SVM)~\cite{cortes1995support} and Random Forest (RF)~\cite{breiman2001random} into cloud detection tasks.

The features used in traditional machine learning methods mainly include brightness, texture, and local statistical features. Brightness features are based on low-level pixel-wise spectral reflectance; for example, Kang \emph{et al.} proposed an unsupervised method~\cite{kang2018coarse} that employs SVM to segment clouds in the HSI space, while Fu \emph{et al.} introduced Random Forest into cloud detection~\cite{fu2018cloud} to enhance the modeling of nonlinear feature relationships. Texture features provide higher-level semantic information; for instance, Chen \emph{et al.}~\cite{chen2007support} extracted texture features using the Gray-Level Co-occurrence Matrix (GLCM) and combined them with a nonlinear SVM for cloud segmentation, and Sui \emph{et al.}~\cite{sui2019energy} utilized SLIC superpixels and Gabor responses to derive texture representations. Local statistical features further enhance discriminative capability through distribution modeling; for example, Yuan \emph{et al.}~\cite{yuan2015bag} employed a Bag-of-Words model to extract statistical features, Tan \emph{et al.}~\cite{tan2016cloud} integrated multiple spectral and structural features, and Deng \emph{et al.}~\cite{deng2018cloud} further introduced natural scene statistics features. Such methods improve feature utilization efficiency, but their performance remains limited by feature representation and scene generalization capability.

\subsection{Deep Learning-Based Cloud Detection in Remote Sensing Images}
Due to the limited representation capacity of handcrafted features, traditional machine learning methods suffer from performance degradation in complex scenarios~\cite{wu2021geographic}. Since 2012, deep learning models such as CNNs have achieved remarkable progress in computer vision tasks, including image classification and object detection, and have been gradually introduced into the remote sensing domain.

Early deep learning-based approaches formulated cloud detection as a binary classification problem on image patches. Mateo \emph{et al.}~\cite{mateo2017convolutional} divided images into $33\times33$ patches and applied CNNs for classification. Xie \emph{et al.}~\cite{xie2017multilevel} utilized SLIC to segment images into superpixels instead of direct patch cropping. However, since image patches may contain both cloudy and clear pixels, such methods are prone to classification errors. Inspired by Fully Convolutional Networks (FCNs)~\cite{long2015fully}, deep learning-based cloud detection gradually evolved toward pixel-wise dense prediction~\cite{wu2018utilizing,yan2018cloud}. The U-Net architecture~\cite{ronneberger2015u}, with its symmetric structure and skip connections, significantly improved segmentation performance. Numerous cloud detection methods based on U-Net have been proposed~\cite{francis2019cloudfcn,wieland2019multi,jeppesen2019cloud}, substantially advancing detection accuracy.

To address challenging scenarios such as thin-cloud ambiguity, irregular boundaries, bright snow, and building interference, various specialized models have been proposed. Yang \emph{et al.}~\cite{yang2019cdnet} employed feature pyramids and edge refinement modules to improve detection accuracy in low-resolution imagery. Yu \emph{et al.}~\cite{yu2020effective} designed a dual-branch CNN to extract shallow and deep features, incorporating pyramid pooling and spatial attention for enhanced feature fusion. Li \emph{et al.}~\cite{li2020deep} leveraged satellite physical imaging mechanisms to guide fine cloud detection. Wu \emph{et al.}~\cite{wu2021geographic} incorporated geographic information such as latitude, longitude, and elevation for accurate cloud-snow separation. Yang \emph{et al.}~\cite{yang2024weakly} proposed a weakly supervised cloud-snow detection framework to suppress snow interference.

In recent years, the introduction of attention mechanisms has further improved the performance of cloud detection models. Zhang \emph{et al.} proposed CAA-UNet~\cite{zhang2021improving}, which integrates residual connections and attention mechanisms into the U-Net architecture to enhance cloud feature preservation. Guo \emph{et al.} proposed Cloud-AttU~\cite{guo2020cloud}, introducing attention modules to learn more effective cloud features and improve robustness against interference. With the rise of Transformer in the field of computer vision, related methods have gradually been extended to cloud detection tasks. Liu \emph{et al.} proposed TransCloudSeg~\cite{liu2022transcloudseg}, which constructs a dual-path architecture combining CNN and Transformer to fuse local and global features. Zhang \emph{et al.} proposed Cloudformer~\cite{zhang2022cloudformer}, which employs a dual-decoder structure based on CNN and Transformer to accurately classify similar objects. Building upon this, Zhang \emph{et al.} further proposed the lightweight CloudViT model~\cite{zhang2022cloudvit}, which improves cross-sensor cloud detection performance through multi-scale dark channel guidance. Ge \emph{et al.} proposed CD-CTFM~\cite{ge2024cd}, designing a lightweight CNN-Transformer backbone to achieve a balance between accuracy and computational efficiency. Gong \emph{et al.} proposed the STCCD model~\cite{gong2023hybrid}, which incorporates Swin Transformer to construct a comprehensive representation of clouds through feature fusion and multi-scale aggregation. These works demonstrate that the introduction of attention mechanisms and Transformer architectures can effectively enhance the modeling capability of complex cloud patterns and significantly improve cloud detection performance.

Although CNN- and Transformer-based cloud detection methods have achieved significant progress, they still exhibit certain limitations in complex remote sensing scenarios. First, CNNs are constrained by their local receptive fields and struggle to fully capture global contextual information; although Transformers can model long-range dependencies via self-attention mechanisms, their quadratic computational complexity leads to computational and memory bottlenecks when processing high-resolution remote sensing imagery. To address this issue, this paper adopts a hybrid CNN-Mamba representation architecture, which extracts discriminative local features while efficiently modeling long-range dependencies among features with linear computational complexity.
Second, most existing methods formulate cloud detection as a single-stage forward pixel-wise binary classification problem. This paradigm tends to produce ambiguities when handling semi-transparent thin clouds and complex cloud boundaries, and lacks mechanisms for uncertainty estimation and refinement. To this end, this paper proposes the CloudMamba model, which introduces an uncertainty-guided two-stage refinement framework. Specifically, the second stage performs targeted optimization on low-confidence regions identified by the uncertainty estimation module, thereby effectively alleviating segmentation ambiguity and improving model robustness.

\subsection{Mamba}
Recently, State Space Models (SSMs)~\cite{gu2021efficiently,gu2021combining} have rapidly advanced in the vision domain. Representative models such as Mamba achieve an effective balance between long-range dependency modeling and linear computational complexity~\cite{gu2023mamba}. Compared with CNNs that focus on local receptive fields and Transformers that rely on global self-attention mechanisms, Mamba employs dynamic selection mechanisms to efficiently model spatiotemporal structures and contextual relationships, emerging as a new paradigm beyond CNNs and Transformers. Vision Mamba models have been widely applied to image classification, semantic segmentation, detection, and reconstruction tasks~\cite{he2024igroupss,zhao2024rs,zhang2024cdmamba,liu2024rscama}.

In remote sensing, particularly in segmentation tasks, Mamba has demonstrated promising potential. RS$^3$Mamba~\cite{ma2024rs} directly applied visual state space models to remote sensing scene segmentation, validating their effectiveness in large-scale texture and structural representation. CM-UNet~\cite{liu2024cm} and UNet-Mamba~\cite{zhu2024unetmamba} adopted CNN-Mamba hybrid encoder-decoder structures, leveraging CNNs for local texture representation and Mamba for long-range spatial dependency modeling, achieving improved boundary continuity and regional consistency. PyramidMamba~\cite{wang2025pyramidmamba} and PPMamba~\cite{hu2024ppmamba} further incorporated Mamba into multi-scale feature fusion frameworks to enhance cross-scale semantic interaction, making them suitable for complex remote sensing scenarios with significant object scale variations. These studies demonstrate that Mamba can enhance collaborative modeling of large-scale contextual relationships and fine-grained boundary details in remote sensing segmentation while maintaining efficient inference and low computational overhead, paving the way for its application in cloud detection and change detection tasks.

Although state space models have demonstrated strong potential in remote sensing image segmentation tasks, existing methods typically adopt general-purpose SSM modules, which remain suboptimal for task-specific scenarios such as cloud detection. To address the scale differences between clouds and ground objects as well as the multi-scale characteristics of clouds, this paper designs a dual-scale Mamba block (DS-Mamba) tailored for remote sensing imagery. This module employs a dual-branch collaborative modeling strategy: on the one hand, it captures the global structural information of clouds at a large scale to achieve accurate cloud recognition; on the other hand, it models fine-grained features of cloud boundaries and thin-cloud regions at a small scale, thereby enabling precise cloud segmentation.

\section{Method}
In this section, we present the proposed CloudMamba model for remote sensing image cloud detection based on dual-scale Mamba with uncertainty-guided refinement. The framework is designed for multispectral satellite imagery and adopts an encoder-decoder architecture with an uncertainty-guided refinement network, forming a two-stage deep neural network. CloudMamba integrates convolution and Mamba hybrid modules to extract high-quality local image features while efficiently modeling long-range dependencies.
The proposed dual-scale Mamba module captures macro-level cloud structures at a large scale while preserving fine-grained details at a small scale, thereby enabling accurate cloud recognition and precise delineation of cloud boundaries and thin-cloud regions.
An additionally designed cascaded second-stage refinement module enhances the low-confidence regions of the first-stage mask under the guidance of the uncertainty estimation map, thereby improving the segmentation quality of the final predicted mask.

\begin{figure*}
\centering
\includegraphics[width=1.0\linewidth]{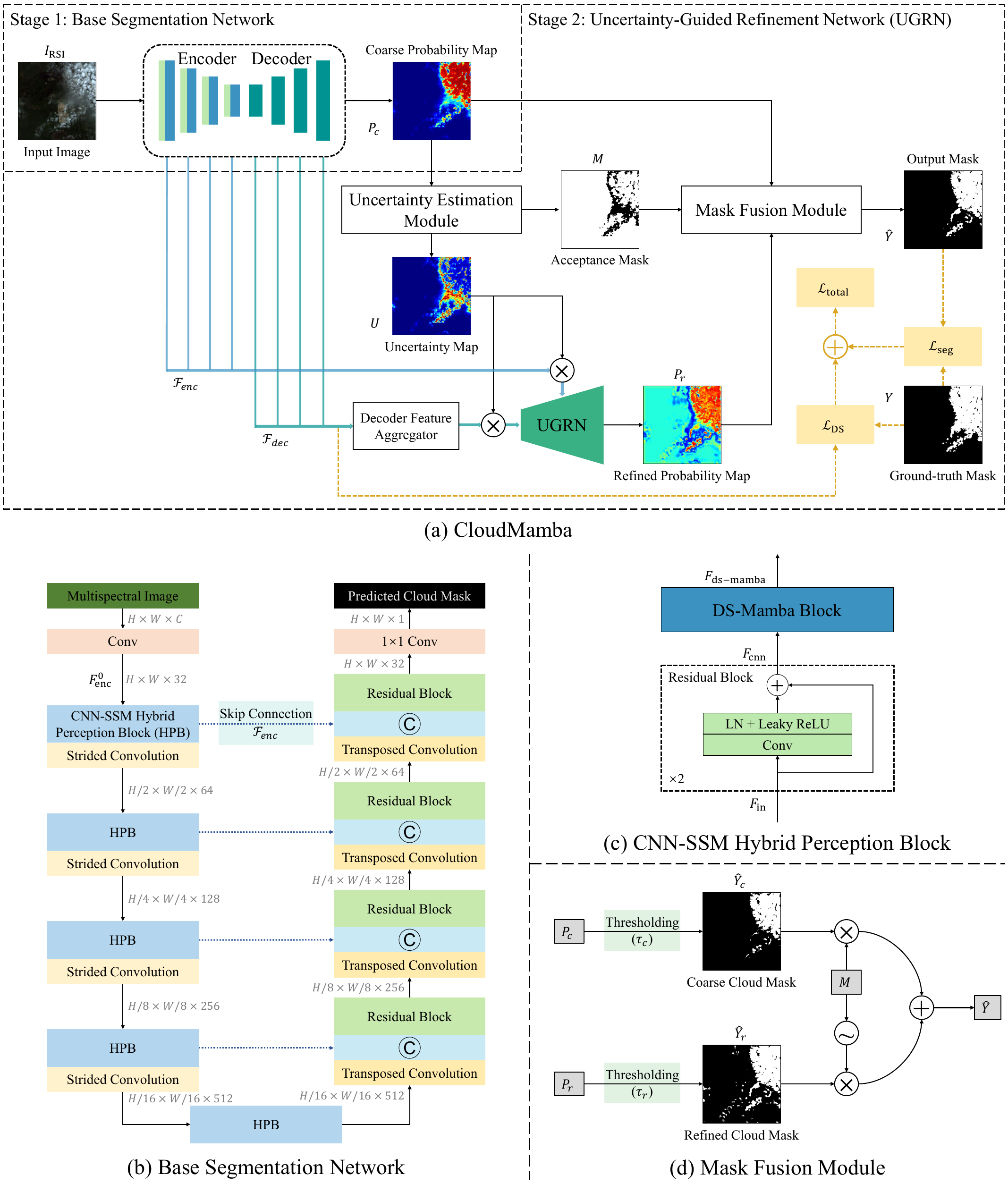}
\caption{Overall architecture of CloudMamba. (a) The complete framework consisting of a base segmentation network generating a coarse prediction mask and a second-stage uncertainty-guided refinement network. (b) The encoder-decoder architecture of the base segmentation network. (c) The CNN-SSM Hybrid Perception Block. (d) The mask fusion module.}\label{fig:network_overall}
\end{figure*}

As shown in Fig.~\ref{fig:network_overall}, the input remote sensing image is fed into the encoder, where multi-scale image features are extracted through stacked CNN-State Space Model Hybrid Perception Blocks (HPBs) along with downsampling layers.
The Dual-Scale Mamba (DS-Mamba) block integrates two branches to jointly exploit small-scale local textures and large-scale macro-structural information, producing more discriminative feature representations.
The hierarchical features extracted by the encoder are passed through the decoder to predict a coarse cloud probability map, from which a coarse cloud mask and its corresponding uncertainty estimation map are computed.
Predictions in low-uncertainty regions are directly accepted, whereas high-uncertainty regions are refined by the Uncertainty-Guided Refinement Network (UGRN), which leverages fused features from both the encoder and decoder under guidance of the uncertainty map.
The overall architecture adopts a cascaded two-stage enhanced segmentation framework in an encoder-decoder-refiner configuration, employing convolutional and Mamba layers to model local features and global context, respectively.
The DS-Mamba block effectively improves the representation capability of the features.
Through the extraction of discriminative features and hierarchical processing, the model's accuracy and robustness for cloud detection in complex scenarios are significantly enhanced.
% two-stage、cascaded
% encoder level、block

Next, we provide a detailed description of each component of the network. The organization of this section is as follows:
Section 3.1 overviews the overall network architecture.
Section 3.2 presents the dual-scale Mamba block and the CNN-SSM hybrid perception module built upon it.
Section 3.3 introduces the base segmentation network with an encoder-decoder architecture.
Section 3.4 describes the uncertainty-guided refinement network, including uncertainty map estimation and final cloud mask generation.
Section 3.5 discusses the loss functions.

\subsection{Overview of the CloudMamba Architecture}
As shown in Fig.~\ref{fig:network_overall}(a), the overall CloudMamba framework consists of a base segmentation network with an encoder-decoder architecture and a second-stage uncertainty-guided refinement network.
The input to the network is a multispectral remote sensing image $I_\mathrm{RSI}\in\mathbb{R}^{H\times W\times C}$, where $H$ and $W$ denote the height and width of the input image in pixels, respectively, and $C$ represents the number of spectral bands.
First, discriminative multi-scale features $\mathcal{F}_\mathrm{enc}$ are extracted by the encoder. The decoder then reconstructs a coarse cloud probability map $P_c\in[0,1]^{H\times W}$ based on these features.
Subsequently, an uncertainty estimation map $U\in[0,1]^{H\times W}$ and an acceptance mask $M\in\{0,1\}^{H\times W}$ for the base segmentation predictions are computed from the coarse cloud probability map, where $1$ indicates that the prediction is accepted and $0$ indicates that further determination by the refinement network is required.
In the second stage, the encoder and decoder features are first modulated using the uncertainty estimation map. These modulated features are then fed into the refinement network for enhanced segmentation to obtain a refined cloud probability map $P_r\in[0,1]^{H\times W}$.
Finally, the coarse and refined cloud probability maps are fused pixel-by-pixel, guided by the acceptance mask, to generate the final cloud mask prediction. CloudMamba comprises the following key submodules:
% ground-truth mask: $Y\in\{0,1\}^{H\times W}$
% coarse segmentation mask: $\hat{Y}_c=1(P_c>\tau_c)$
% refined segmentation mask: $\hat{Y}_r=1(P_r>\tau_r)$
% final cloud segmentation result mask: $\hat{Y}$

\textbf{Encoder.} The input remote sensing image first passes through an initial convolutional layer for preliminary feature extraction, mapping multi-channel image pixels into a high-dimensional feature space. Subsequently, the image features are fed into $L$ consistently stacked encoder levels to extract discriminative multi-scale features:
\begin{equation}
    \mathcal{F}_{\mathrm{enc}}=\{F_{\mathrm{enc}}^1,F_{\mathrm{enc}}^2,\dots,F_{\mathrm{enc}}^L\} ,
\end{equation}
\begin{equation}
    F_{\mathrm{enc}}^l\in\mathbb{R}^{H_l\times W_l\times C_l},\quad l=1,\dots,L ,
\end{equation}
Each encoder level consists of a CNN-SSM hybrid perception module followed by a strided convolution layer, which perform multi-scale feature extraction and feature downsampling, respectively. The HPB module sequentially integrates residual blocks for capturing local detailed features and a dual-scale Mamba block for modeling global contextual dependencies.

\textbf{Decoder.} The lightweight decoder is composed of stacked residual blocks and progressively reconstructs high-level cloud semantic features $\mathcal{F}_{\mathrm{enc}}$ from the discriminative multi-scale features $\mathcal{F}_{\mathrm{dec}}$ extracted by the encoder, ultimately generating a high-precision cloud mask. Each decoder level employs a transposed convolution layer to upsample feature maps and gradually restore the spatial resolution of the input remote sensing image.
In addition, skip connections are used to fuse high-level semantic features from the decoder with high-resolution detailed features from the encoder to improve segmentation accuracy.
Finally, a $1\times1$ convolution layer is applied to produce the coarse cloud mask prediction of the first stage.

\textbf{Refiner.} The second-stage uncertainty-guided refinement network further enhances the segmentation of "low-confidence" regions in the coarse cloud mask predicted by the base segmentation network.
The UGRN adopts the same network architecture as the decoder of the base segmentation network.
First, the uncertainty estimation map is used to modulate the feature maps $\mathcal{F}_{\mathrm{enc}},\mathcal{F}_{\mathrm{dec}}$ from the base segmentation network.
Then, the modulated decoder features are fused with the encoder features to reconstruct the cloud prediction mask for low-confidence regions in the coarse cloud mask.
Finally, guided by the acceptance mask, the coarse cloud mask from the first stage and the refined cloud mask from the second stage are fused pixel-by-pixel to generate the final output cloud mask.

\subsection{CNN-SSM Hybrid Perception Module}
The inherent inductive bias of convolutional neural networks naturally aligns with the structural characteristics of image data. CNN layers can efficiently extract hierarchical features from low-level details to high-level abstract semantics. However, due to the local receptive field of convolution operations, CNNs have difficulty in effectively modeling long-range global contextual dependencies. Mamba leverages state space models (SSMs) to efficiently capture global dependencies in sequential data, compensating for the inherent limitations of CNNs while maintaining high computational efficiency. The CNN-SSM hybrid perception module adopts a hybrid design that combines convolution and state space modeling, enabling the extraction of high-quality local image features while efficiently modeling long-range dependencies between features. This design allows the network to learn more discriminative remote sensing features in complex scenarios such as fragmented thin clouds and cloud-like ground objects.

\subsubsection{Mamba Architecture}
State Space Models (SSMs)~\cite{gu2023modeling} are used to model the temporal evolution of one-dimensional functions or sequences $u(t)\in\mathbb{R}\mapsto y(t)\in\mathbb{R}$, which can be formulated as the following linear ordinary differential equations (ODEs):
\begin{equation}
\begin{aligned}
    x^{\prime}(t) &= \mathbf{A}x(t)+\mathbf{B}u(t) ,\\
    y(t) &= \mathbf{C}x(t) ,
\end{aligned}
\end{equation}
where $\mathbf{A}\in\mathbb{R}^{N\times N},\mathbf{B},\mathbf{C}\in\mathbb{R}^N$ are model parameters, and $x(t)\in\mathbb{R}^N$ denotes the implicit latent state.
% implicit latent state

Structured State Space Sequence Models (S4)~\cite{gu2022efficiently} impose structured constraints on the state matrix $\mathbf{A}$ and introduce efficient algorithms, thereby improving stability and computational efficiency.
The selective Structured State Space Sequence Model, Mamba (S6)~\cite{gu2023mamba}, further introduces an input-dependent selection mechanism, enabling the model to dynamically and efficiently perform information filtering.

\subsubsection{Dual-Scale Mamba Block}
Mamba overcomes the limitation of the local receptive field in traditional CNNs and the quadratic computational complexity bottleneck of Transformer models, enabling efficient modeling of global dependencies in input data. Considering the multi-scale characteristics of remote sensing imagery, on the one hand, clouds and ground objects exhibit significant spatial scale variations, and single-scale features are insufficient to reliably represent both. On the other hand, the high variability in cloud scales introduces considerable uncertainty, making it difficult for single-scale features to stably extract discriminative cloud representations. Consequently, Mamba layers based on fixed-resolution features struggle to accurately model remote sensing image characteristics. To address this issue, we design a dual-scale Mamba block tailored for remote sensing imagery, which can accurately identify clouds at large structural scales while finely processing cloud boundaries and thin cloud regions at small detail scales. Through the collaboration of dual-branch features, the modeling capability of the Mamba block for remote sensing information is enhanced.

\begin{figure}
\centering
\includegraphics[width=1.0\linewidth]{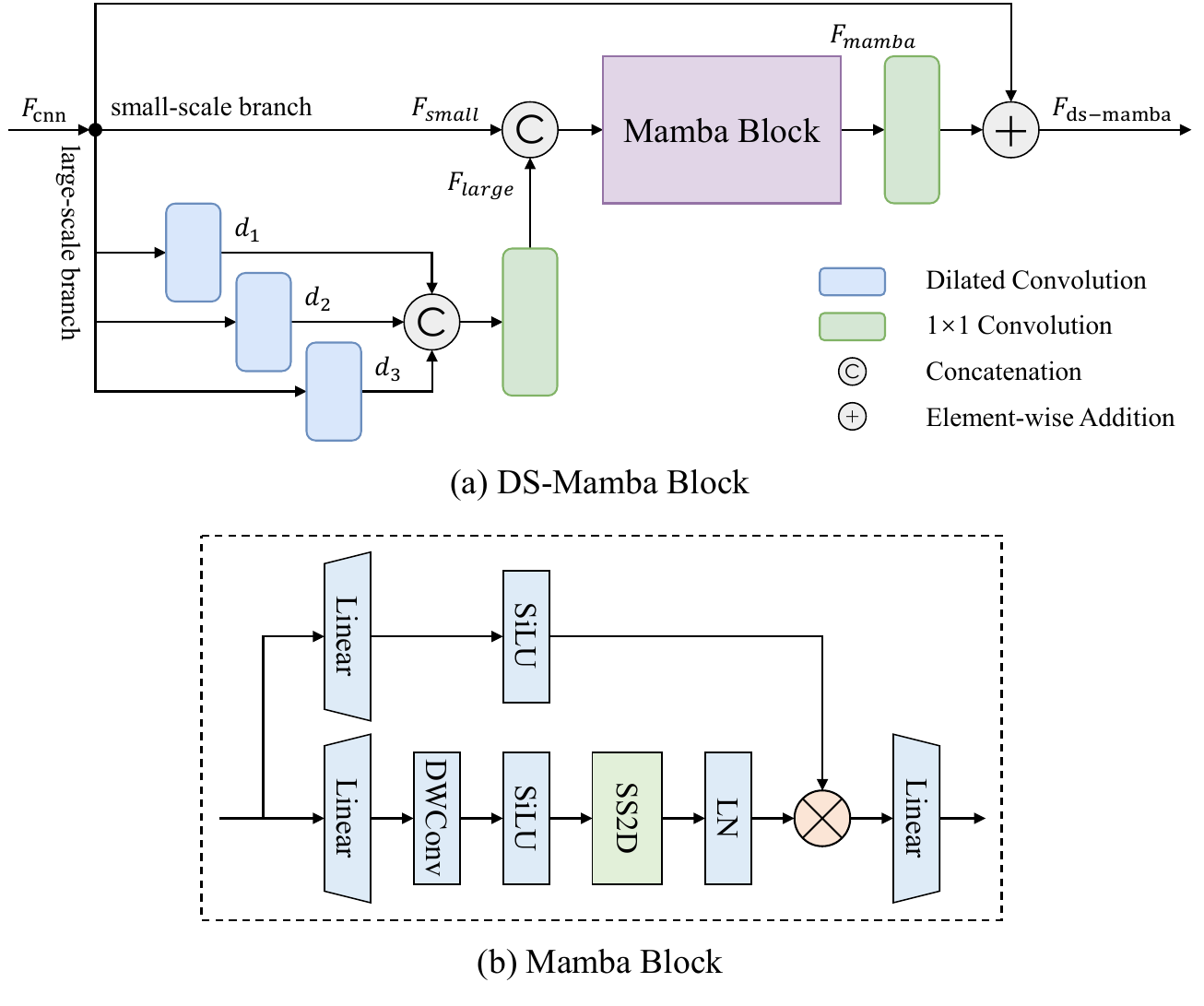}
\caption{Network structure of the dual-scale Mamba (DS-Mamba) block. (a) Structure of the DS-Mamba block. (b) Structure of the Mamba block~\cite{liu2024vmamba}.}\label{fig:network_dsmamba}
\end{figure}

The network structure of the DS-Mamba block is shown in Fig.~\ref{fig:network_dsmamba}(a). To naturally extend the Mamba architecture, originally designed for one-dimensional sequential data, to two-dimensional image modeling, we follow the best practices of existing studies by replacing the SSM block for sequence transformation in the original Mamba with a 2D Selective Scan (SS2D) block~\cite{liu2024vmamba}, which performs image scanning in four directions (left-to-right, top-to-bottom, right-to-left, and bottom-to-top), thereby effectively capturing global contextual information within the two-dimensional image space.
As illustrated in Fig.~\ref{fig:network_dsmamba}(a), the input feature $F_{\mathrm{cnn}}$ is fed in parallel into a small-scale feature branch and a large-scale feature branch. The large-scale branch employs three parallel dilated convolution layers (with dilation rates $d_1,d_2,d_3$) to enlarge the receptive field of the input features. The resulting feature maps are concatenated along the channel dimension, followed by a $1\times1$ convolutional layer for channel dimension reduction:
\begin{equation}
    F_{\mathrm{small}}=F_{\mathrm{cnn}} ,
\end{equation}
\begin{equation}
\begin{split}
    F_{\mathrm{large}} = &\mathrm{Conv}_{1\times1}\Big(\mathrm{Concat}\big(
    \mathrm{Conv}_{d_1}(F_{\mathrm{cnn}}),\\
    &\mathrm{Conv}_{d_2}(F_{\mathrm{cnn}}),
    \mathrm{Conv}_{d_3}(F_{\mathrm{cnn}})
    \big)\Big) ,
\end{split}
\end{equation}
where $F_{\mathrm{small}}$ and $F_{\mathrm{large}}$ denote the output feature maps of the small-scale and large-scale branches, respectively. $\mathrm{Conv}_{d_i}$ represents a dilated convolution operation with dilation rate $d_i$, and the $1\times1$ convolution layer $\mathrm{Conv}_{1\times1}$ reduces the channel number of the concatenated features to match that of $F_{\mathrm{small}}$.
Subsequently, the dual-scale features from the two branches are concatenated and fed into the Mamba block for feature mapping:
\begin{equation}
    F_{\mathrm{mamba}} = \mathrm{Mamba}\big(\mathrm{Concat}(F_{\mathrm{small}},
    F_{\mathrm{large}})\big) ,
\end{equation}
Then, a $1\times1$ convolution layer restores the channel dimension of the feature map, and a residual connection with the input feature is applied to obtain the final output of the DS-Mamba block:
\begin{equation}
    F_{\mathrm{ds-mamba}} = F_{\mathrm{cnn}}+\mathrm{Conv}_{1\times1}(F_{\mathrm{mamba}}) .
\end{equation}

The structure of the Mamba block is shown in Fig.~\ref{fig:network_dsmamba}(b). The layer-normalized input feature is passed into parallel branches for linear projection. The main branch feature first goes through a depthwise convolution (DWConv) layer with a SiLU activation function, followed by the SS2D block and layer normalization. The gated branch feature is activated by SiLU and multiplied element-wise with the output feature of the main branch. Finally, a linear projection is applied to the output feature to obtain the final output of the Mamba block.

\begin{figure*}
\centering
\includegraphics[width=1.0\linewidth]{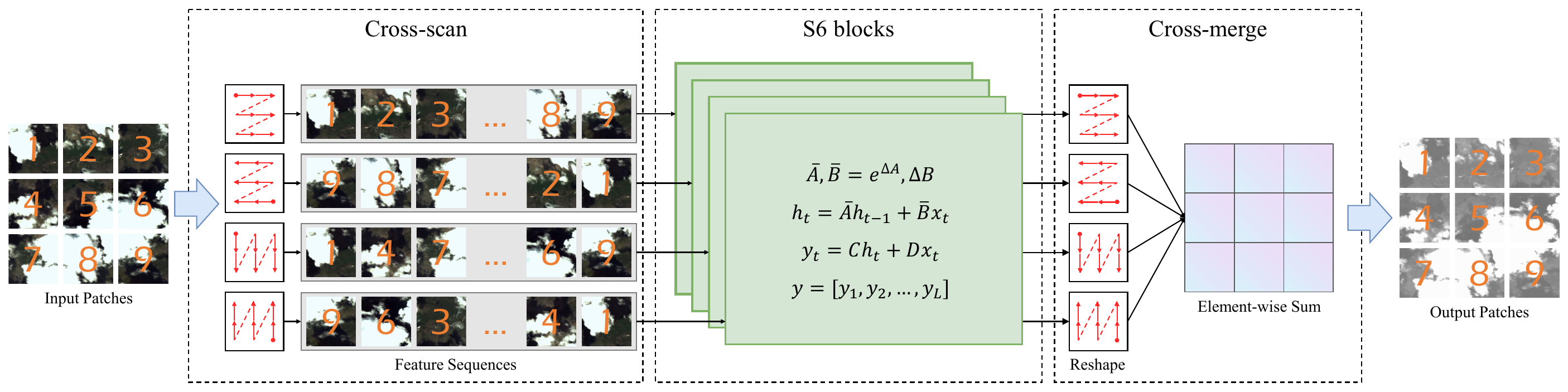}
\caption{Four-directional scanning mechanism of 2D Selective Scan (SS2D).}\label{fig:network_ss2d}
\end{figure*}

As shown in Fig.~\ref{fig:network_ss2d}, the SS2D block consists of three components: cross-scan, selective scan S6 blocks, and cross-merge. The cross-scan operation unfolds the two-dimensional feature map into one-dimensional sequences along four different directions (enabling the modeling of contextual dependencies in four directions: left-to-right, top-to-bottom, right-to-left, and bottom-to-top).
Correspondingly, the cross-merge operation reshapes the output sequences from different unfolding directions and fuses them via element-wise summation to reconstruct the two-dimensional feature map. The internal selective scan component applies the S6 block to each one-dimensional sequence along each direction to perform sequence mapping, enabling efficient feature modeling and information filtering. Through feature modeling in four directions, the network can comprehensively perceive the contextual dependencies between clouds and ground objects from different spatial perspectives, thereby obtaining robust global feature representations to cope with complex remote sensing scenarios.
% reshape

\subsubsection{CNN-SSM Hybrid Perception Module}
The network structure of the CNN-SSM hybrid perception module is shown in Fig.~\ref{fig:network_dsmamba}(c). The input feature first passes through two serially stacked convolutional residual blocks for local feature extraction:
\begin{equation}
    F_{\mathrm{cnn}} = \mathrm{ResBlock}_2\big(\mathrm{ResBlock}_1(F_{\mathrm{in}})\big) ,
\end{equation}
where each $\mathrm{ResBlock}(\cdot)$ consists of a convolution layer, layer normalization (LN), and a Leaky ReLU activation function. The preceding convolutional residual blocks efficiently extract high-quality local image features such as edges and texture patterns at different network levels.
Subsequently, the features extracted by the residual blocks are fed into a dual-scale Mamba block for global contextual modeling:
\begin{equation}
    F_{\mathrm{ds-mamba}} = \mathrm{DSMamba}(F_{\mathrm{cnn}}) ,
\end{equation}
The DS-Mamba block efficiently models long-range dependencies between features. Its dedicated dual-scale design facilitates the synthesis of more discriminative feature representations, enabling the network to distinguish clouds from interfering ground objects using large-scale features while achieving precise segmentation of cloud boundaries and thin cloud regions using small-scale details.

\subsection{Base Segmentation Network}
As shown in Fig.~\ref{fig:network_dsmamba}(b), the base segmentation network adopts an encoder-decoder architecture. The input remote sensing image $I_\mathrm{RSI}$ first passes through an initial convolution layer to extract preliminary image features:
\begin{equation}
    F_{\mathrm{enc}}^0 = \mathrm{Conv}(I_\mathrm{RSI}) ,
\end{equation}
Then, the feature map $F_{\mathrm{enc}}^0$ is sequentially fed into $L$ CNN-SSM hybrid perception modules with strided convolutions for downsampling, progressively extracting multi-scale discriminative features:
\begin{equation}
    F_{\mathrm{enc}}^l = \mathrm{Conv}_{s=2}\big(\mathrm{HPB}(F_{\mathrm{enc}}^{l-1})\big),\quad l=1,\dots,L ,
\end{equation}
where $\mathrm{HPB}(\cdot)$ denotes the CNN-SSM hybrid perception module, $\mathrm{Conv}_{s=2}(\cdot)$ represents a convolution operation with stride $2$, and $F_{\mathrm{enc}}^l$ denotes the output feature of the $l$-th encoder level.

The decoder consists of $L$ levels corresponding to the encoder and aims to progressively reconstruct the cloud coverage mask by accurately localizing cloud pixels leveraging the discriminative high-level semantic feature $F_{\mathrm{enc}}^L$ extracted by the encoder. At each decoder level, a transposed convolution operation is first used to upsample the low-resolution feature map. Then, through skip connections, the high-resolution detailed feature $\mathcal{F}_\mathrm{enc}$ from the corresponding encoder level is concatenated with the high-level semantic feature $\mathcal{F}_\mathrm{dec}$ from the decoder, leveraging low-level spatial details to improve segmentation accuracy at cloud boundaries and thin cloud regions. The concatenated feature is fed into a lightweight convolutional module composed of two cascaded residual blocks for deep feature fusion and fine-grained enhancement. This process can be formulated as:
\begin{equation}
\begin{split}
    F_\mathrm{dec}^l = &\mathrm{ResBlock}_2\bigg(\mathrm{ResBlock}_1\Big(\mathrm{Concat} \\
    &\big(\mathrm{TransConv}(F_\mathrm{dec}^{l+1}),F_\mathrm{enc}^l\big)\Big)\bigg),\quad l=L,\dots,1 ,
\end{split}
\end{equation}
where $\mathrm{TransConv}(\cdot)$ denotes the transposed convolution upsampling layer, $\mathrm{Concat}(\cdot)$ represents channel-wise concatenation of feature maps, $\mathrm{ResBlock}_i(\cdot)$ denotes the $i$-th residual block, and $F_\mathrm{dec}^l$ is the output feature map of the $l$-th decoder level.
Finally, the high-level semantic feature $F_\mathrm{dec}^1$ output by the decoder is projected through a $1\times1$ convolution layer to generate the coarse cloud mask probability map of the first stage:
\begin{equation}
    P_c = \sigma\big(\mathrm{Conv}_{1\times1}(F_\mathrm{dec}^1)\big) ,
\end{equation}
where $\mathrm{Conv}_{1\times1}(\cdot)$ denotes a $1\times1$ convolution operation, and $\sigma(\cdot)$ is the Sigmoid activation function that maps the output into the $[0,1]$ cloud pixel probability space.

\subsection{Uncertainty-Guided Refinement Network}
\subsubsection{Uncertainty Map Estimation}
To further improve the prediction accuracy of the base segmentation network in complex regions such as thin clouds and cloud boundaries, the second-stage uncertainty-guided refinement network focuses on low-confidence regions in the coarse cloud mask generated in the first stage and performs targeted enhanced segmentation and fine correction on these regions, thereby improving the overall cloud detection performance of the model.
First, the coarse cloud mask probability map $P_c$ generated in the first stage is fed into the uncertainty estimation module to obtain the uncertainty map corresponding to the coarse cloud mask $\hat{Y}_c$. The specific computation is as follows:
\begin{equation}
    U = 1-2\left|P_c-0.5\right| ,
\end{equation}
where $U\in[0,1]^{H\times W}$ represents the prediction uncertainty at each pixel location of $\hat{Y}_c$.
It is worth noting that the uncertainty map is not directly supervised during training, but is deterministically computed from the coarse probability map $P_c$ without introducing additional learnable parameters.
This design allows the uncertainty to naturally reflect the confidence of the base segmentation network predictions, where probabilities close to $0.5$ indicate higher ambiguity, while avoiding the need for explicit uncertainty labels.

Then, the uncertainty estimation map $U$ is utilized to calculate the acceptance mask $M\in\{0,1\}^{H\times W}$ for the prediction results of the base segmentation network:
\begin{equation}
    M = \mathds{1}(U<\gamma) ,
\end{equation}
where $\mathds{1}(\cdot)$ denotes the indicator function, and $\gamma\in(0,1)$ represents the uncertainty threshold. Predictions at locations with uncertainty lower than this threshold are accepted, while the remaining regions are passed to the refinement network for further determination.

\subsubsection{Final Cloud Mask Generation}
\textbf{Refined cloud probability prediction.} To fully exploit the multi-scale semantic information from the decoder of the base segmentation network, the output features $\{F_{\mathrm{dec}}^l\}_{l=1}^L$ from all decoder levels are fed into a decoder feature aggregator for fusion. Specifically, the decoder features at each level are first resized to the same spatial resolution as the decoder input feature map via bilinear interpolation. They are then concatenated along the channel dimension and compressed through a $1\times1$ convolution layer for channel reduction and feature remapping, yielding the aggregated decoder feature:
\begin{equation}
\begin{split}
    F_{\mathrm{dec}}^{\mathrm{agg}} = &\mathrm{Conv}_{1\times1}\Big(\mathrm{Concat} \\
    &\big(\mathrm{Down}(F_{\mathrm{dec}}^1),\dots,\mathrm{Down}(F_{\mathrm{dec}}^L)\big)\Big) .
\end{split}
\end{equation}
where $\mathrm{Down}(\cdot)$ denotes bilinear interpolation downsampling.
Subsequently, the uncertainty estimation map $U$ is used to modulate the encoder features $\mathcal{F}_\mathrm{enc}$ from the base segmentation network and the aggregated decoder feature $F_{\mathrm{dec}}^{\mathrm{agg}}$, highlighting discriminative information in low-confidence regions while suppressing redundant features in high-confidence regions. The uncertainty modulation is defined as:
\begin{align}
    \tilde{F}_{\mathrm{enc}}^l &= U \odot F_{\mathrm{enc}}^l,\quad l=1,\dots,L , \\
    \tilde{F}_{\mathrm{dec}}   &= U \odot F_{\mathrm{dec}}^{\mathrm{agg}} ,
\end{align}
where $\odot$ denotes element-wise multiplication, and the uncertainty map $U$ is resized via bilinear interpolation to match the spatial resolution of the corresponding feature map.

The modulated decoder feature $\tilde{F}_{\mathrm{dec}}$ is fed into the second-stage refinement network UGRN, which adopts the same network architecture as the decoder of the base segmentation network. The modulated encoder features $\{\tilde{F}_{\mathrm{enc}}^l\}_{l=1}^L$ are incorporated via skip connections to assist the refiner in fine reconstruction of high-uncertainty regions.
Finally, the high-resolution semantic feature $F_{\mathrm{ref}}$ output by the refiner is projected through a $1\times1$ convolution layer followed by a Sigmoid activation function to generate the refined cloud probability map:
\begin{equation}
    P_r = \sigma\big(\mathrm{Conv}_{1\times1}(F_{\mathrm{ref}})\big) ,
\end{equation}
where $P_r\in[0,1]^{H\times W}$ denotes the refined cloud probability map.

\textbf{Two-stage cloud mask fusion.} First, the coarse cloud mask $\hat{Y}_c$ and the refined cloud mask $\hat{Y}_r$ are obtained from the coarse cloud probability map $P_c$ and the refined cloud probability map $P_r$ via thresholding:
\begin{align}
    \hat{Y}_c &= \mathds{1}(P_c>\tau_c) , \\
    \hat{Y}_r &= \mathds{1}(P_r>\tau_r) ,
\end{align}
where $\tau_c,\tau_r\in(0,1)$ denote the cloud pixel classification thresholds for the base segmentation network and the second-stage refinement network, respectively. $\mathds{1}(\cdot)$ represents the indicator function, which outputs 1 when the cloud probability at a pixel exceeds the threshold and 0 otherwise.
Subsequently, according to the guidance of the acceptance mask $M$, the two-stage prediction masks are fused pixel-wise to generate the final predicted cloud mask $\hat{Y}$:
\begin{equation}
    \hat{Y} = M \odot \hat{Y}_c + (1-M) \odot \hat{Y}_r .
\end{equation}

\subsection{Loss Functions}
Considering that cloud detection typically suffers from class imbalance and that cloud boundaries and thin cloud regions exhibit high spatial uncertainty, we adopt a composite loss function combining Binary Cross-Entropy (BCE) loss and Dice loss, and introduce a deep supervision mechanism to enhance convergence stability and segmentation accuracy across multi-scale features.

Let the cloud probability prediction map output by the network be denoted as $P\in[0,1]^{H\times W}$, and the corresponding ground-truth cloud label be $Y\in\{0,1\}^{H\times W}$. The pixel-wise binary cross-entropy loss is defined as:
\begin{equation}
\begin{split}
    \mathcal{L}_{\mathrm{BCE}} = -\frac{1}{HW}\sum_{i=1}^{H}\sum_{j=1}^{W}\Big[Y_{ij}\log P_{ij} + \\
    (1-Y_{ij})\log(1-P_{ij})\Big] .
\end{split}
\end{equation}
To alleviate the class imbalance problem and encourage more continuous and complete cloud region segmentation, the Dice loss is introduced:
\begin{equation}
    \mathcal{L}_{\mathrm{Dice}} = 1 - \frac{2\sum_{i,j}P_{ij}Y_{ij}+\epsilon}{\sum_{i,j}P_{ij}+\sum_{i,j}Y_{ij}+\epsilon} ,
\end{equation}
where $\epsilon$ is a smoothing term to avoid numerical instability.
By combining the above two losses, the BCE-Dice composite segmentation loss is computed as:
\begin{equation}
    \mathcal{L}_{\mathrm{seg}} = \lambda_{\mathrm{bce}}\mathcal{L}_{\mathrm{BCE}} + \lambda_{\mathrm{dice}}\mathcal{L}_{\mathrm{Dice}} ,
\end{equation}
where $\lambda_{\mathrm{bce}},\lambda_{\mathrm{dice}}$ denote the weighting coefficients for the BCE and Dice loss, respectively.

To stabilize the training process and promote effective learning of multi-scale semantic features, multi-scale prediction heads are introduced in the base segmentation network for deep supervision.
The auxiliary prediction probability maps generated from decoder features $\mathcal{F}_{\mathrm{dec}}$ at different level are denoted as $\{P^l\}_{l=1}^L$.
The ground-truth label $Y$ is downsampled to the corresponding spatial resolution via nearest-neighbor interpolation, denoted as $\{Y^l\}_{l=1}^L$. The deep supervision loss is computed as:
\begin{equation}
    \mathcal{L}_{\mathrm{DS}} = \sum_{l=1}^L \alpha_l\mathcal{L}_{\mathrm{seg}}(P^l, Y^l) ,
\end{equation}
where $\{\alpha_l\}_{l=1}^L$ denote the weighting coefficients for supervision at different scales.

Finally, the total loss function used for network training consists of the composite segmentation loss and the deep supervision loss:
\begin{equation}
    \mathcal{L}_{\mathrm{total}} = \mathcal{L}_{\mathrm{seg}} + \mathcal{L}_{\mathrm{DS}} .
\end{equation}

\section{Experimental Results and Analyses}
\subsection{Experimental Setup}
\subsubsection{Datasets}
To comprehensively evaluate the proposed CloudMamba model, experiments were conducted on two publicly available cloud detection datasets, GF1\_WHU and Levir\_CS.

The GF1\_WHU dataset~\cite{li2017multi} consists of 108 scenes of Level-2A imagery captured by the Wide Field of View (WFV) sensor onboard the Gaofen-1 (GF-1) satellite, along with corresponding reference masks for clouds and cloud shadows. The GF-1 WFV imagery has a spatial resolution of 16 meters and contains four multispectral bands ranging from visible to near-infrared wavelengths. The images were acquired from May 2013 to August 2016, covering diverse global land cover types with varying cloud conditions. The reference masks were manually delineated by experienced annotators to outline cloud and cloud-shadow boundaries. Among the 108 scenes, 86 are used for training and the remaining 22 for testing.

The Levir\_CS dataset~\cite{wu2021geographic} contains 4,168 scenes of GF-1 WFV imagery. The pixel-wise reference masks are manually annotated into three categories: cloud, snow, and background. The dataset covers a global distribution with acquisition dates ranging from May 2013 to February 2019, encompassing diverse surface types such as plains, plateaus, water bodies, deserts, and ice fields. The images include various climatic conditions worldwide, such as desert and marine climates. In the Levir\_CS dataset, all images are downsampled by a factor of 10. Each image has a size of $1320\times1200$ pixels, with a spatial resolution of 160 meters. The dataset is randomly divided into two subsets: 3,068 scenes for training and 1,100 scenes for testing.

In our experiments, we utilized all four spectral bands from the GF1\_WHU and Levir\_CS datasets, and each scene was offline cropped into image patches of size $512\times512$ pixels. After preprocessing, the GF1\_WHU dataset contains 52,826 training images and 13,937 testing images; the Levir\_CS dataset contains 27,612 training images and 9,900 testing images. The reference mask corresponding to each image patch was uniformly defined into two categories: "cloud" and "clear". Specifically, cloud shadows in the reference masks of the GF1\_WHU dataset and snow in the reference masks of the Levir\_CS dataset were both categorized as clear background pixels.

\subsubsection{Implementation Details}
In the experiments, the number of stages $L$ in the base segmentation network is set to 5. The dilation rates $d_1,d_2,d_3$ of the large-scale branch in the DS-Mamba module are set to $1,2,4$, respectively. The cloud pixel classification thresholds $\tau_c$ and $\tau_r$ for the coarse and refined cloud masks are both set to 0.5, and the uncertainty threshold $\gamma$ for accepting predictions from the base segmentation network is set to 0.4.
When computing the BCE-Dice composite segmentation loss, the weighting coefficients $\lambda_{\mathrm{bce}}$ and $\lambda_{\mathrm{dice}}$ for the BCE and Dice loss are both set to 1.

Our model is implemented based on the PyTorch framework and trained on a single NVIDIA GeForce RTX 4090 GPU. Data augmentation techniques applied to the input patches include random flipping and random rotation. During optimization, the AdamW optimizer is employed with an initial learning rate of $1e-4$. A cosine annealing learning rate scheduler (CosineAnnealingLR)~\cite{loshchilov2016sgdr} is adopted to decay the learning rate, and the total number of training epochs is set to 30.

\subsubsection{Evaluation Metrics}
To quantitatively evaluate the performance of cloud detection models, three widely used metrics are adopted: mean Intersection over Union (mIoU), F1-score (F1), and Overall Accuracy (OA).
mIoU measures the regional overlap between the predicted mask and the ground-truth annotation. For the binary classification task, mIoU is computed as:
\begin{equation}
    \mathrm{mIoU} = \frac{1}{2}\left(\frac{\mathrm{TP}}{\mathrm{TP}+\mathrm{FP}+\mathrm{FN}} + \frac{\mathrm{TN}}{\mathrm{TN}+\mathrm{FP}+\mathrm{FN}}\right) ,
\end{equation}
where $\mathrm{TP}$, $\mathrm{TN}$, $\mathrm{FP}$, and $\mathrm{FN}$ denote the numbers of true positive, true negative, false positive, and false negative pixels, respectively.
The F1-score integrates Precision and Recall, and can effectively reflect model performance under class imbalance conditions. Its calculation formula is:
\begin{equation}
    \mathrm{F1} = \frac{2\times\mathrm{TP}}{2\times\mathrm{TP}+\mathrm{FP}+\mathrm{FN}} .
\end{equation}
OA measures the overall prediction accuracy across all pixels and is defined as:
\begin{equation}
    \mathrm{OA} = \frac{\mathrm{TP}+\mathrm{TN}}{\mathrm{TP}+\mathrm{TN}+\mathrm{FP}+\mathrm{FN}} .
\end{equation}

\subsection{Comparison with State-of-the-Art Methods}
We compare the proposed method with several classical and state-of-the-art cloud detection approaches, including U-Net~\cite{ronneberger2015u}, DeepLabV3+~\cite{chen2018encoder}, DCNet~\cite{liu2021dcnet}, BoundaryNets~\cite{wu2022cloud}, CloudU-Net~\cite{shi2020cloudu}, CloudSegNet~\cite{dev2019cloudsegnet}, U-Mamba~\cite{ma2024u}, and VM-UNet~\cite{ruan2024vm}. Among them, U-Net~\cite{ronneberger2015u} is a classical encoder-decoder architecture that achieves fine-grained pixel-level segmentation by fusing low-level and high-level features through skip connections. DeepLabV3+~\cite{chen2018encoder} combines atrous spatial pyramid pooling (ASPP) with an efficient decoder structure to effectively model multi-scale contextual information and refine boundaries.
DCNet~\cite{liu2021dcnet} introduces deformable convolutions in the encoding stage to enhance the model's adaptive capability for modeling cloud shape variations and complex underlying surfaces. BoundaryNets~\cite{wu2022cloud} improves the representation of cloud region boundaries by leveraging multi-scale boundary modeling and differential boundary learning modules. CloudU-Net~\cite{shi2020cloudu} incorporates atrous convolutions and fully connected conditional random field (CRF) post-processing to enhance contextual awareness and segmentation accuracy. CloudSegNet~\cite{dev2019cloudsegnet} adopts a purely convolutional encoder-decoder architecture, enabling unified processing of all-weather cloud segmentation tasks. U-Mamba~\cite{ma2024u} enhances long-range dependency modeling by integrating convolutional operations with state space models (SSMs). VM-UNet~\cite{ruan2024vm} constructs a U-shaped architecture based on pure visual Mamba modules, achieving excellent global context modeling with linear complexity. We implement the above models using their publicly available codes with default hyperparameters.

\begin{table*}
\caption{Comparison results on the GF1\_WHU and Levir\_CS cloud detection test sets. The best scores are highlighted in bold} \label{tab:performance}
\centering
\begin{tabular}{c|ccc|ccc}
	\toprule
	\multirow{2}{*}{Methods} & \multicolumn{3}{c|}{\textbf{GF1\_WHU}} & \multicolumn{3}{c}{\textbf{Levir\_CS}}\\
	& mIoU (\%) $\uparrow$ & F1 (\%) $\uparrow$ & OA (\%) $\uparrow$ & mIoU (\%) $\uparrow$ & F1 (\%) $\uparrow$ & OA (\%) $\uparrow$\\
	\midrule
	U-Net~\cite{ronneberger2015u}            & 84.72 & 91.73 & 95.46 & 89.43 & 94.42 & 97.96\\
	DeepLabV3+~\cite{chen2018encoder}        & 86.66 & 92.85 & 95.91 & 81.10 & 89.56 & 96.10\\
    DCNet~\cite{liu2021dcnet}                & 84.07 & 91.35 & 95.23 & 71.70 & 83.52 & 94.24\\
    BoundaryNets~\cite{wu2022cloud}          & 87.86 & 93.54 & 96.32 & 90.34 & 94.92 & 98.12\\
    CloudU-Net~\cite{shi2020cloudu}          & 84.47 & 91.58 & 95.32 & 86.27 & 92.63 & 97.23\\
    CloudSegNet~\cite{dev2019cloudsegnet}    & 82.51 & 90.42 & 94.71 & 64.84 & 78.67 & 92.29\\
    U-Mamba~\cite{ma2024u}                   & 86.94 & 93.01 & 95.96 & 90.78 & 95.17 & 98.22\\
    VM-UNet~\cite{ruan2024vm}                & 72.74 & 84.22 & 91.29 & 82.76 & 90.56 & 96.51\\
    CloudMamba & \textbf{89.27} & \textbf{94.33} & \textbf{96.78} & \textbf{91.24} & \textbf{95.42} & \textbf{98.31}\\
	\bottomrule
\end{tabular}
\end{table*}

Table~\ref{tab:performance} presents the comprehensive comparison results on the GF1\_WHU and Levir\_CS test sets. The quantitative results demonstrate that the proposed CloudMamba model significantly outperforms all competing methods across all evaluation metrics on both datasets.
On the GF1\_WHU dataset, compared with the second-best method BoundaryNets, our CloudMamba achieves improvements of approximately 1.4, 0.8, and 0.5 percentage points in terms of mIoU, F1, and OA, respectively.
On the Levir\_CS dataset, compared with the best-performing competitor U-Mamba, the proposed model improves mIoU, F1, and OA by approximately 0.5, 0.3, and 0.1 percentage points, respectively. These results indicate that CloudMamba exhibits more stable segmentation performance and stronger generalization capability in complex remote sensing scenarios, validating the effectiveness of the proposed dual-scale Mamba module for multi-scale feature representation and global contextual modeling, as well as the advantage of the uncertainty-guided refinement framework in adaptively enhancing segmentation in challenging image regions.

\subsection{Qualitative Comparisons}

\begin{figure*}
\centering
\includegraphics[width=1.0\linewidth]{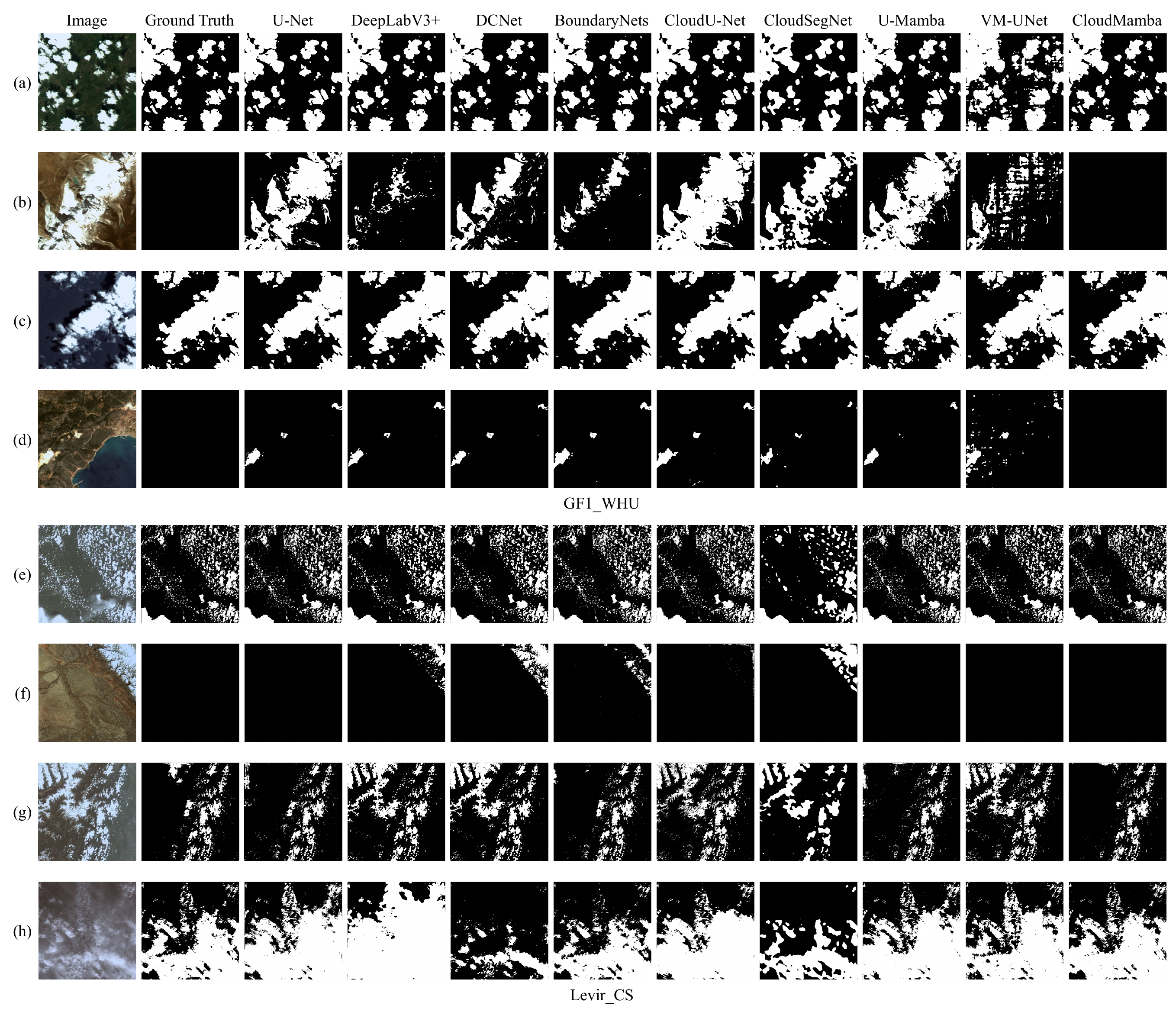}
\caption{Visualization results of different methods on the GF1\_WHU and Levir\_CS test sets. (a--h) represent the prediction results of different methods on various samples.}\label{fig:comparison}
\end{figure*}

The visual comparison results of different methods on the two datasets are shown in Fig.~\ref{fig:comparison}. It can be observed that, compared with other approaches, CloudMamba achieves superior segmentation performance under various complex scenarios. First, for certain ground objects that are highly similar to clouds in appearance, CloudMamba can more effectively avoid false detections (as shown in Fig.~\ref{fig:comparison}(b), (d), (f), and (g)).
For example, in Fig.~\ref{fig:comparison}(b) and (f), most competing methods misclassify bright snow-covered regions partially or entirely as clouds. Benefiting from the global contextual modeling capability of the DS-Mamba module, CloudMamba learns more discriminative feature representations, thereby effectively distinguishing clouds from snow and significantly reducing false positives. In Fig.~\ref{fig:comparison}(d), due to the high similarity in color characteristics and structural patterns between bright ground objects and clouds, most comparison methods produce misclassifications, whereas CloudMamba more accurately suppresses such interference regions. In the complex scenario shown in Fig.~\ref{fig:comparison}(g), where ridge snow and clouds coexist with highly overlapping spatial distributions, most methods fail to effectively differentiate them, while CloudMamba can still accurately identify cloud-covered areas. Second, benefiting from the introduction of the uncertainty-guided refinement strategy, CloudMamba produces more precise segmentation results in challenging regions such as cloud boundaries. As illustrated in Fig.~\ref{fig:comparison}(a), (c), and (e), some methods (e.g., CloudSegNet) generate cloud masks with coarse boundaries, whereas CloudMamba outputs masks with clearer boundaries, more complete structures, and fewer missed detections in fragmented cloud regions. Finally, in thin cloud scenarios (as shown in Fig.~\ref{fig:comparison}(h)), CloudMamba effectively balances the discrimination between clouds and background objects, reducing both missed detections and false alarms in thin cloud areas and achieving more accurate and reliable segmentation results.

\subsection{Ablation Study}
\subsubsection{Ablation of Proposed Components}

\begin{table*}
\caption{Ablation study of key components. The best scores are highlighted in bold} \label{tab:ablation_component}
\centering
\begin{tabular}{cccc|ccc}
	\toprule
    Mamba & DS-Mamba (Sep) & DS-Mamba & UGRN & mIoU (\%) $\uparrow$ & F1 (\%) $\uparrow$ & OA (\%) $\uparrow$\\
	\midrule
    \ding{55} & \ding{55} & \ding{55} & \ding{55} & 87.71 & 93.45 & 96.31\\
	\ding{51} & \ding{55} & \ding{55} & \ding{55} & 88.81 & 94.07 & 96.65\\
    \ding{55} & \ding{51} & \ding{55} & \ding{55} & 89.09 & 94.23 & 96.70\\
    \ding{55} & \ding{55} & \ding{51} & \ding{55} & 89.21 & 94.30 & 96.75\\
	\ding{55} & \ding{55} & \ding{51} & \ding{51} & \textbf{89.27} & \textbf{94.33} & \textbf{96.78}\\
	\bottomrule
\end{tabular}
\end{table*}
% (1) Base[Encoder(CNN)-Decoder]
% (2) 1+Mamba
% (3) 1+DS-Mamba(Separated)
% (4) 1+DS-Mamba
% (5) 3+UGRN[Full]

To verify the contribution of each proposed key component to the overall performance, ablation experiments are conducted on the GF1\_WHU dataset, and the results are shown in Table~\ref{tab:ablation_component}. First, when using only a pure convolutional encoder-decoder architecture as the baseline model, the network achieves 87.71, 93.45, and 96.31 in terms of mIoU, F1, and OA, respectively.
On this basis, introducing the standard Mamba module to enhance global modeling capability leads to significant improvements across all metrics, demonstrating the effectiveness of state space model-based global dependency modeling for cloud detection. Furthermore, replacing the standard Mamba with the proposed DS-Mamba module yields further performance gains compared to the single-scale Mamba, indicating that the dual-scale design can more adequately characterize cloud patterns at different spatial scales in remote sensing imagery, thereby producing more discriminative feature representations.
Finally, after incorporating the uncertainty-guided refinement network (UGRN), the model achieves the best performance, reaching 89.27, 94.33, and 96.78 in mIoU, F1, and OA, respectively. This result confirms that the uncertainty-guided refinement strategy effectively focuses on low-confidence regions in the base segmentation results and performs targeted enhancement on challenging areas such as cloud boundaries and thin clouds, thereby further improving overall segmentation accuracy.

Additionally, to investigate the optimal fusion mechanism for dual-branch features, we evaluate a variant of DS-Mamba, termed DS-Mamba (Sep). In this variant, the dual-branch features are first processed in parallel by their respective DS-Mamba modules, followed by feature fusion. As shown in Table~\ref{tab:ablation_component}, DS-Mamba (Sep) performs worse than the standard DS-Mamba across all evaluation metrics. This result indicates that performing feature fusion at an earlier stage facilitates more effective cross-scale feature interaction within the Mamba module, thereby enabling the learning of more discriminative cloud representations at different spatial resolutions and fully leveraging the advantage of SSM in global context modeling.

\subsubsection{Ablation on Dilation Rates in the Large-Scale Branch}

\begin{table}
\caption{Ablation study of dilation rate configurations in the large-scale branch. All scores are reported in percentage (\%)} \label{tab:ablation_dilation}
\centering
\begin{tabular}{ccc|ccc}
	\toprule
    $d_1$ & $d_2$ & $d_3$ & mIoU $\uparrow$ & F1 $\uparrow$ & OA $\uparrow$\\
	\midrule
    1 & 1 & 1 & 88.87 & 94.11 & 96.67\\
	2 & 2 & 2 & 88.96 & 94.16 & 96.72\\
	1 & 2 & 4 & \textbf{89.27} & \textbf{94.33} & \textbf{96.78}\\
    2 & 4 & 8 & 89.02 & 94.19 & 96.70\\
	\bottomrule
\end{tabular}
\end{table}
% (1) [1,1,1]
% (2) [2,2,2]
% (3) [1,2,4](Model)
% (4) [2,4,8]

To investigate the effect of the receptive field size of the large-scale branch in the dual-scale Mamba module on model performance, we conduct ablation experiments on the dilation rates $d_1,d_2,d_3$ of three groups of dilated convolution layers, and the results are presented in Table~\ref{tab:ablation_dilation}.
Specifically, four configurations are compared: (1) a baseline configuration without multi-scale design ($\{1,1,1\}$); (2) a single-scale configuration with fixed dilation rates ($\{2,2,2\}$); (3) a multi-scale configuration with progressively increasing dilation rates ($\{1,2,4\}$); and (4) a large receptive field configuration with larger dilation rates ($\{2,4,8\}$). When the three dilated convolution layers share identical dilation rates ($\{1,1,1\}$ or $\{2,2,2\}$), the model performance is relatively limited, indicating that a single receptive field scale is insufficient to fully capture the multi-scale characteristics of clouds and ground objects in remote sensing imagery.
In contrast, adopting the progressively increasing dilation rate configuration $\{1,2,4\}$ yields the best performance in terms of mIoU, F1, and OA, indicating that the combination of multi-scale receptive fields can more effectively integrate local details and macro-structural information, thereby enhancing the overall discrimination and fine-grained segmentation capability for clouds.
Further increasing the dilation rates to $\{2,4,8\}$ results in a slight performance degradation, which may be attributed to the introduction of redundant background information from excessively large receptive fields, weakening the model's ability to finely characterize cloud features.
Considering both accuracy and stability, we finally adopt $\{1,2,4\}$ as the default dilation rate configuration for the large-scale branch.

\subsection{Comparison Between Single-Stage and Two-Stage Frameworks}

\begin{table}
\caption{Comparison results between single-stage and two-stage frameworks. All scores are reported in percentage (\%)} \label{tab:single_two_stage}
\centering
\begin{tabular}{c|ccc}
    \toprule
    Frameworks & mIoU $\uparrow$ & F1 $\uparrow$ & OA $\uparrow$\\
    \midrule
    Single-stage (w/o UGRN) & 89.21 & 94.30 & 96.75\\
    Two-stage (CloudMamba) & \textbf{89.27} & \textbf{94.33} & \textbf{96.78}\\
    \bottomrule
\end{tabular}
\end{table}

\begin{figure}
\centering
\includegraphics[width=1.0\linewidth]{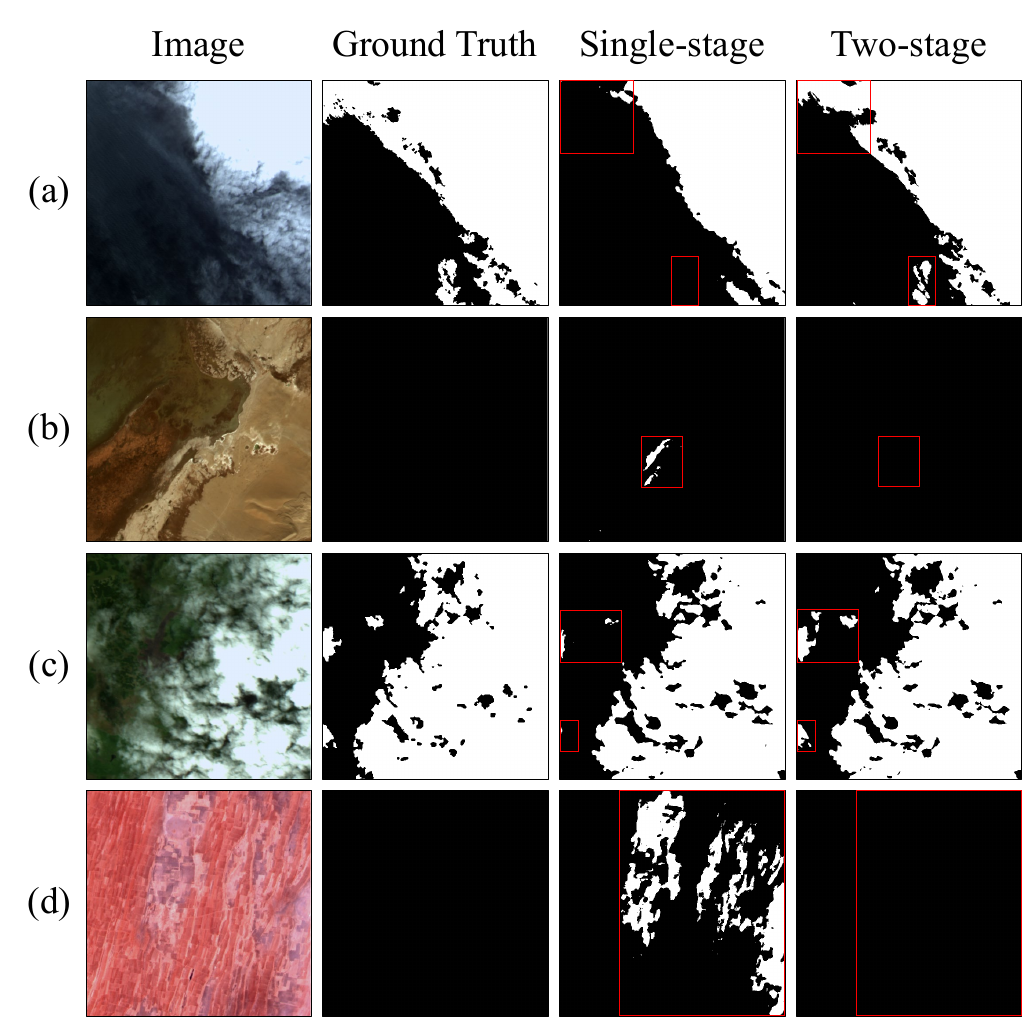}
\caption{Visualization of prediction results from single-stage and two-stage frameworks on the test set. The test samples are from the GF1\_WHU dataset.}\label{fig:single_two_stage}
\end{figure}

To further validate the effectiveness of the proposed uncertainty-guided two-stage refinement framework, the complete CloudMamba model is compared with its corresponding single-stage variant. The single-stage model retains only the base segmentation network in the CloudMamba architecture, removing the uncertainty-guided refinement network.
The quantitative evaluation results on the GF1\_WHU test set are presented in Table~\ref{tab:single_two_stage}. With the integration of the second-stage refinement network, the model exhibits improvements across mIoU, F1, and OA metrics.
Although the overall performance gain is modest, the two-stage framework with a lightweight pure convolutional refiner effectively improves prediction quality in challenging regions such as cloud boundaries and thin clouds while maintaining controllable computational complexity. Fig.~\ref{fig:single_two_stage} presents visual comparisons of prediction masks on representative samples.
As observed from the red rectangular regions in Fig.~\ref{fig:single_two_stage}(a), the single-stage model exhibits obvious missed detections of thin clouds.
It is also evident in Fig.~\ref{fig:single_two_stage}(c) that the two-stage CloudMamba model produces more complete segmentation of thin cloud regions with clearer boundaries.
From Fig.~\ref{fig:single_two_stage}(b) and (d), it can be observed that the single-stage model tends to misclassify complex textures and high-brightness regions of ground objects as clouds, whereas the two-stage model avoids such false alarms.
The comparison results with the single-stage model demonstrate the effectiveness and necessity of the proposed uncertainty-guided two-stage refinement strategy.

\subsection{Performance on Hard Samples}
To further validate the effectiveness of the proposed uncertainty-guided two-stage refinement framework in challenging scenarios such as thin clouds, fragmented clouds, and complex cloud boundaries, a hard-sample subset is constructed from the GF1\_WHU test set. Quantitative comparisons between the single-stage and two-stage frameworks are conducted on this subset. Specifically, based on the uncertainty estimation maps produced by the base segmentation network, the average prediction uncertainty is computed for each test image, and the top $10\%$ samples with the highest uncertainty scores are selected to form the hard-sample subset.

\begin{table}
\caption{Performance comparison on the hard-sample subset. All metrics are reported in percentage (\%)} \label{tab:hard_subset}
\centering
\begin{tabular}{c|ccc}
    \toprule
    Methods & mIoU $\uparrow$ & F1 $\uparrow$ & OA $\uparrow$\\
    \midrule
    Single-stage CloudMamba & 73.21 & 84.53 & 80.70\\
    Two-stage CloudMamba (Ours) & \textbf{74.85} & \textbf{85.61} & \textbf{81.94}\\
    \bottomrule
\end{tabular}
\end{table}

Table~\ref{tab:hard_subset} presents the quantitative comparison results between the single-stage CloudMamba and the two-stage CloudMamba with the uncertainty-guided refinement network on the hard-sample subset. It can be observed that the two-stage model significantly outperforms the single-stage model in terms of mIoU, F1, and OA, indicating that the proposed uncertainty-guided refinement strategy can effectively focus on low-confidence prediction regions and perform targeted enhanced segmentation for challenging areas such as cloud boundaries and thin clouds in hard samples, thereby improving the model's segmentation accuracy and robustness in complex scenarios.

\subsection{Qualitative Analysis of the CloudMamba Framework}

\begin{figure*}
\centering
\includegraphics[width=1.0\linewidth]{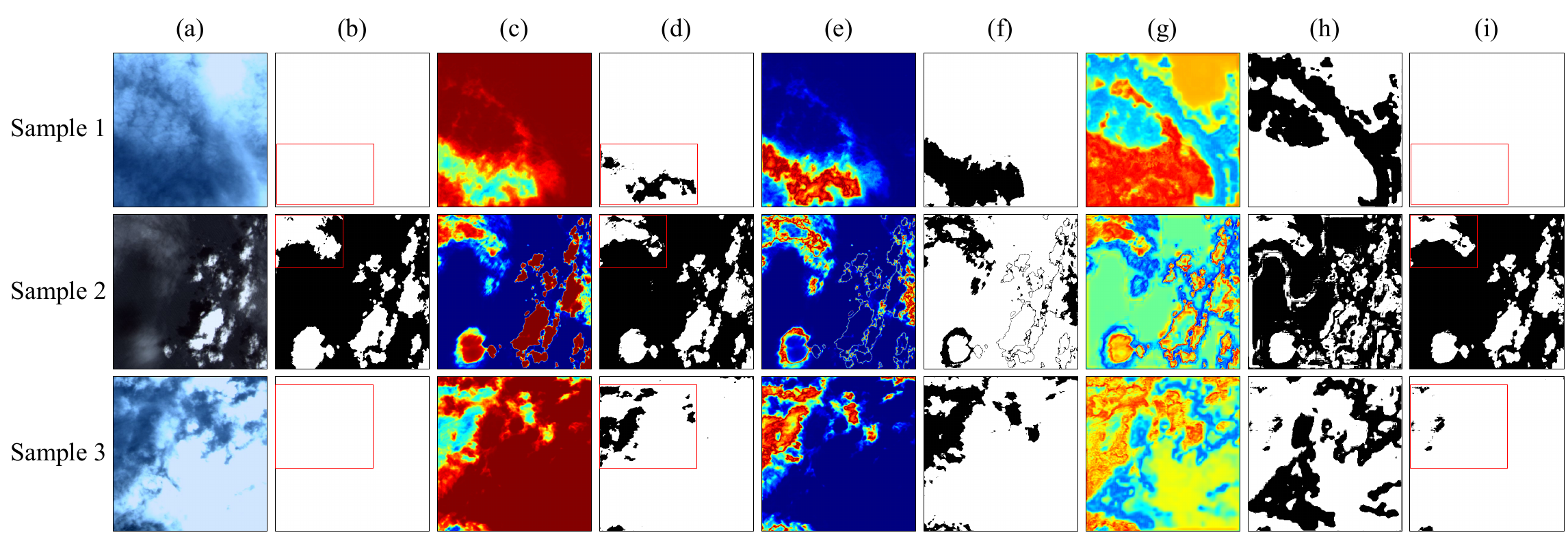}
\caption{Visualization examples of different stages in the CloudMamba framework. (a) Input image $I_\mathrm{RSI}$, (b) ground-truth cloud mask $Y$, (c) coarse cloud probability map $P_c$, (d) coarse cloud mask $\hat{Y}_c$, (e) uncertainty estimation map $U$, (f) acceptance mask $M$, (g) refined cloud probability map $P_r$, (h) refined cloud mask $\hat{Y}_r$, and (i) output cloud mask $\hat{Y}$. The test samples are from the GF1\_WHU dataset.}\label{fig:visualization}
\end{figure*}
% Input Image, Ground-Truth Mask, Coarse Probability Map, Coarse Cloud Mask, Uncertainty Map, Acceptance Mask, Refined Probability Map, Refined Cloud Mask, Output Mask.

To better demonstrate the effectiveness and underlying mechanism of the proposed uncertainty-guided refinement framework, we visualize and analyze the key intermediate feature maps and prediction results of the CloudMamba network.
As shown in Fig.~\ref{fig:visualization}, three representative test samples are presented, including the coarse cloud probability maps and masks, uncertainty estimation maps and acceptance masks, refined prediction results, and final output cloud masks.
% Sample 1
It can be observed from the region marked by the red rectangle in Sample 1(d) that the coarse cloud mask output by the first-stage base segmentation network suffers from missed detections in some low-brightness image regions. Sample 1(e) further reveals high uncertainty in this region, whereas the refined cloud mask obtained through the second-stage enhanced segmentation (Sample 1(h)) successfully compensates for the missed detections in this region (Sample 1(i)).
% Sample 3
The coarse cloud mask of Sample 3(d) exhibits a cloud omission issue similar to that of Sample 1, while the refined and enhanced cloud mask (Sample 3(i)) significantly reduces the area of missed detections, effectively improving cloud detection accuracy.
% Sample 2
In the mixed scenario of thick and thin clouds in Sample 2, the coarse cloud mask (the region marked by the red rectangle in Sample 2(d)) produces inaccurate segmentation results in the thin cloud area, whereas the refined cloud mask (Sample 2(i)) achieves a more complete segmentation of this thin cloud region with clearer boundaries.
% Summary
These visualization results further validate the effectiveness of the CloudMamba framework in leveraging uncertainty maps to guide the second-stage enhanced segmentation. This mechanism effectively improves the model's cloud detection accuracy in complex scenarios such as cloud boundaries and thin clouds.

\subsection{Discussion}
The proposed CloudMamba model demonstrates superior performance in remote sensing cloud detection tasks, and its performance gains on the hard-sample subset (see Table~\ref{tab:hard_subset}) further validate its effectiveness in complex scenarios.
Nevertheless, several limitations remain and warrant further investigation. First, although the uncertainty-guided refinement network (UGRN) can significantly improve segmentation accuracy in challenging regions such as thin clouds and cloud boundaries, the adopted cascaded two-stage architecture inevitably introduces additional computational overhead and memory consumption compared to single-stage models. Despite the lightweight design of the refinement network, this additional cost may still limit its applicability in scenarios with strict real-time requirements or constrained computational resources. Second, the current uncertainty map is computed heuristically based on the distance between the predicted probability and the decision boundary ($0.5$). While this strategy can efficiently reflect the confidence of first-stage predictions, it may not be sufficient to model complex epistemic and aleatoric uncertainties, such as those caused by sensor noise or out-of-distribution shifts under specific imaging conditions.
Finally, although the DS-Mamba module effectively enhances multi-scale feature representation capability, its structural design still relies on predefined dilation rate configurations, which may not always achieve optimal adaptation when encountering complex cloud distributions and spatial patterns.

Future research can be conducted along several directions. First, exploring the incorporation of adaptive or learnable multi-scale modeling mechanisms within the Mamba framework could further enhance its ability to represent complex spatial patterns. Second, more advanced and learnable uncertainty quantification methods, such as Bayesian neural networks or evidential deep learning, can be introduced to provide more reliable uncertainty guidance. Third, more lightweight or dynamic refinement strategies can be designed, for example, by selectively activating the second stage through an adaptive triggering mechanism, thereby reducing computational overhead while maintaining performance. Finally, extending the dual-scale Mamba architecture to other remote sensing tasks (e.g., change detection) is also a promising research direction. Given the advantage of DS-Mamba in capturing scale-varying targets, it is expected to play an important role in modeling spatial dynamics in such tasks.

\section{Conclusion}
In this paper, we propose CloudMamba, a novel two-stage cloud detection framework for remote sensing imagery.
Specifically, a base segmentation network with an encoder-decoder architecture is first constructed, within which a dual-scale Mamba module is introduced.
By efficient collaborative modeling of large-scale macro-structural features and small-scale local detail features, the model's discriminative capability for challenging regions such as thin clouds, fragmented clouds, and cloud boundaries is significantly enhanced.
Building upon this, an uncertainty-guided refinement segmentation mechanism is further designed. By explicitly modeling the pixel-level uncertainty of the first-stage predictions, low-confidence regions are adaptively localized, and the second-stage refinement network is guided to perform targeted enhanced segmentation, effectively improving the overall segmentation accuracy of the model.
Extensive experimental results on the GF1\_WHU and Levir\_CS public remote sensing cloud detection datasets demonstrate that the proposed CloudMamba framework outperforms existing mainstream methods across evaluation metrics such as mIoU, F1, and OA, and exhibits stronger robustness and accuracy stability, particularly in scenarios involving thin clouds, cloud boundaries, and complex backgrounds.

{\small
\bibliographystyle{IEEEtran}
\bibliography{references}
}

\end{document}